\pdfoutput=1

\documentclass[11pt]{article}

\usepackage[dvipsnames, svgnames, x11names]{xcolor}
\usepackage{acl}

\newcommand*\samethanks[1][\value{footnote}]{\footnotemark[#1]}
\usepackage{times}
\usepackage{latexsym}
\usepackage{amsfonts}

\usepackage[T1]{fontenc}

\usepackage[utf8]{inputenc}

\usepackage{microtype}

\usepackage{booktabs}
\usepackage{xcolor}
\usepackage{tabularx}
\usepackage{multirow}
\usepackage{amsmath}
\usepackage{amsfonts}
\usepackage{subcaption}
\usepackage{pgfplots}
\usepackage{pgfplotstable}
\usepackage{pifont}
\usepackage{colortbl}
\usepackage{CJK}
\usepackage{arydshln}
\usepackage{booktabs}
\usepackage{multirow}
\usepackage{bbding}
\usepackage{tikz}
\usetikzlibrary {arrows.meta}
\usepackage[normalem]{ulem}
\useunder{\uline}{\ul}{}
\newcommand{\abbr}{DiffusionBERT}

\title{DiffusionBERT: Improving Generative Masked Language Models with Diffusion Models}

\author{Zhengfu He\thanks{\ \ \ Equal contribution.}\quad
Tianxiang Sun\samethanks\quad
Kuanning Wang\quad
Xuanjing Huang\quad
Xipeng Qiu\\
School of Computer Science, Fudan University\\
Shanghai Key Laboratory of Intelligent Information Processing, Fudan University\\
\texttt{\{zfhe19,txsun19,wangkn20,xjhuang,xpqiu\}@fudan.edu.cn}
}

\begin{document}
\maketitle

\begin{abstract}
We present DiffusionBERT, a new generative masked language model based on discrete diffusion models.
Diffusion models and many pre-trained language models have a shared training objective, i.e., \textit{denoising}, making it possible to combine the two powerful models and enjoy the best of both worlds. 
On the one hand, diffusion models offer a promising training strategy that helps improve the generation quality.
On the other hand, pre-trained denoising language models (e.g., BERT) can be used as a good initialization that accelerates convergence.
We explore training BERT to learn the reverse process of a discrete diffusion process with an absorbing state and elucidate several designs to improve it.
First, we propose a new noise schedule for the forward diffusion process that controls the degree of noise added at each step based on the information of each token.
Second, we investigate several designs of incorporating the time step into BERT.
Experiments on unconditional text generation demonstrate that DiffusionBERT achieves significant improvement over existing diffusion models for text (e.g., D3PM and Diffusion-LM) and previous generative masked language models in terms of perplexity and BLEU score.\footnote{Our code is publicly available at \url{https://github.com/Hzfinfdu/Diffusion-BERT}}

\end{abstract}

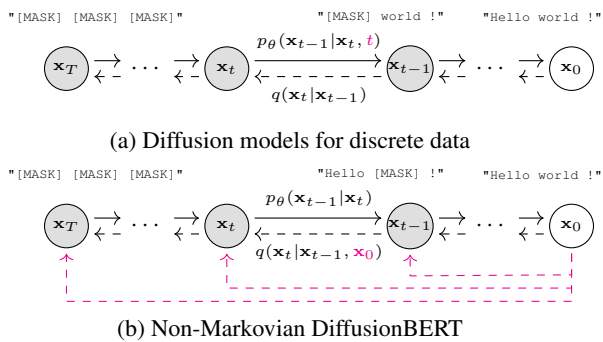
\begin{figure}[t!]
    \centering
    \begin{subfigure}{\linewidth}
    \centering
    \begin{tikzpicture}
    \tikzstyle{gray circle}=[circle,fill=gray!25,minimum size=1.5em,inner sep=0pt,draw=black];
    \tikzstyle{white circle}=[circle,minimum size=1.5em,inner sep=0pt,draw=black];
    
    \node[gray circle] (x_T) at (0,0) {\tiny{$\mathbf{x}_T$}};
    \node (dots1) at ([xshift=2em]x_T.east) {\small{$\cdots$}};
    \node[gray circle] (x_t) at ([xshift=1.8em]dots1.east) {\tiny{$\mathbf{x}_t$}};
    \node[gray circle] (x_t-1) at ([xshift=5.5em]x_t.east) {\tiny{$\mathbf{x}_{t-1}$}};
    \node (dots2) at ([xshift=2em]x_t-1.east) {\small{$\cdots$}};
    \node[white circle] (x_0) at ([xshift=1.8em]dots2.east) {\tiny{$\mathbf{x}_{0}$}};
    \node (noise) at ([xshift=1em,yshift=1em]x_T.north) {\fontsize{5pt}{6pt}{\texttt{"[MASK] [MASK] [MASK]"}}};
    \node (sentt) at ([xshift=-1em,yshift=1em]x_t-1.north) {\fontsize{5pt}{6pt}{\texttt{"[MASK] world !"}}};
    \node (sent0) at ([xshift=-1em,yshift=1em]x_0.north) {\fontsize{5pt}{6pt}{\texttt{"Hello world !"}}};
    
    \draw [->] ([xshift=2pt,yshift=3pt]x_T.east) -- ([yshift=3pt]dots1.west);
    \draw [->,dashed] ([yshift=-3pt]dots1.west) -- ([xshift=2pt,yshift=-3pt]x_T.east);
    
    \draw [->] ([yshift=3pt]dots1.east) -- ([xshift=-2pt,yshift=3pt]x_t.west);
    \draw [->,dashed] ([xshift=-2pt,yshift=-3pt]x_t.west) -- ([yshift=-3pt]dots1.east);
    
    \draw [->] ([xshift=2pt,yshift=3pt]x_t.east) -- node [text width=5em,text centered,midway,above] {\tiny{$p_\theta (\mathbf{x}_{t-1}\vert\mathbf{x}_t, {\color{magenta}{t}})$}} ([xshift=-2pt,yshift=3pt]x_t-1.west);
    \draw [->,dashed] ([xshift=-2pt,yshift=-3pt]x_t-1.west) -- node [text width=5em,text centered,midway,below] {\tiny{$q (\mathbf{x}_t\vert\mathbf{x}_{t-1})$}} ([xshift=2pt,yshift=-3pt]x_t.east);
    
    \draw [->] ([xshift=2pt,yshift=3pt]x_t-1.east) -- ([yshift=3pt]dots2.west);
    \draw [->,dashed] ([yshift=-3pt]dots2.west) -- ([xshift=2pt,yshift=-3pt]x_t-1.east);
    
    \draw [->] ([yshift=3pt]dots2.east) -- ([xshift=-2pt,yshift=3pt]x_0.west);
    \draw [->,dashed] ([xshift=-2pt,yshift=-3pt]x_0.west) -- ([yshift=-3pt]dots2.east);
    \end{tikzpicture}
    \caption{Diffusion models for discrete data}
    \end{subfigure}
    \begin{subfigure}{\linewidth}
    \centering
    \begin{tikzpicture}
    \tikzstyle{gray circle}=[circle,fill=gray!25,minimum size=1.5em,inner sep=0pt,draw=black];
    \tikzstyle{white circle}=[circle,minimum size=1.5em,inner sep=0pt,draw=black];
    
    \node[gray circle] (x_T) at (0,0) {\tiny{$\mathbf{x}_T$}};
    \node (dots1) at ([xshift=2em]x_T.east) {\small{$\cdots$}};
    \node[gray circle] (x_t) at ([xshift=1.8em]dots1.east) {\tiny{$\mathbf{x}_t$}};
    \node[gray circle] (x_t-1) at ([xshift=5.5em]x_t.east) {\tiny{$\mathbf{x}_{t-1}$}};
    \node (dots2) at ([xshift=2em]x_t-1.east) {\small{$\cdots$}};
    \node[white circle] (x_0) at ([xshift=1.8em]dots2.east) {\tiny{$\mathbf{x}_{0}$}};
    \node (noise) at ([xshift=1em,yshift=1em]x_T.north) {\fontsize{5pt}{6pt}{\texttt{"[MASK] [MASK] [MASK]"}}};
    \node (sentt) at ([xshift=-1em,yshift=1em]x_t-1.north) {\fontsize{5pt}{6pt}{\texttt{"Hello [MASK] !"}}};
    \node (sent0) at ([xshift=-1em,yshift=1em]x_0.north) {\fontsize{5pt}{6pt}{\texttt{"Hello world !"}}};
    
    \draw [->] ([xshift=2pt,yshift=3pt]x_T.east) -- ([yshift=3pt]dots1.west);
    \draw [->,dashed] ([yshift=-3pt]dots1.west) -- ([xshift=2pt,yshift=-3pt]x_T.east);
    
    \draw [->] ([yshift=3pt]dots1.east) -- ([xshift=-2pt,yshift=3pt]x_t.west);
    \draw [->,dashed] ([xshift=-2pt,yshift=-3pt]x_t.west) -- ([yshift=-3pt]dots1.east);
    
    \draw [->] ([xshift=2pt,yshift=3pt]x_t.east) -- node [text width=5em,text centered,midway,above] {\tiny{$p_\theta (\mathbf{x}_{t-1}\vert\mathbf{x}_t)$}} ([xshift=-2pt,yshift=3pt]x_t-1.west);
    \draw [->,dashed] ([xshift=-2pt,yshift=-3pt]x_t-1.west) -- node [text width=5em,text centered,midway,below] {\tiny{$q (\mathbf{x}_t\vert\mathbf{x}_{t-1},{\color{magenta}{\mathbf{x}_0}})$}} ([xshift=2pt,yshift=-3pt]x_t.east);
    
    \draw [->] ([xshift=2pt,yshift=3pt]x_t-1.east) -- ([yshift=3pt]dots2.west);
    \draw [->,dashed] ([yshift=-3pt]dots2.west) -- ([xshift=2pt,yshift=-3pt]x_t-1.east);
    
    \draw [->] ([yshift=3pt]dots2.east) -- ([xshift=-2pt,yshift=3pt]x_0.west);
    \draw [->,dashed] ([xshift=-2pt,yshift=-3pt]x_0.west) -- ([yshift=-3pt]dots2.east);
    
    \draw [->,dashed,magenta] ([yshift=-2pt]x_0.south) -- ([yshift=-10pt]x_0.south) -- ([yshift=-10pt]x_t-1.south) -- ([yshift=-2pt]x_t-1.south);
    \draw [->,dashed,magenta] ([yshift=-2pt]x_0.south) -- ([yshift=-15pt]x_0.south) -- ([yshift=-15pt]x_t.south) -- ([yshift=-2pt]x_t.south);
    \draw [->,dashed,magenta] ([yshift=-2pt]x_0.south) -- ([yshift=-20pt]x_0.south) -- ([yshift=-20pt]x_T.south) -- ([yshift=-2pt]x_T.south);
    
    \end{tikzpicture}
    \caption{Non-Markovian~\abbr}
    \end{subfigure}
    \caption{In contrast to conventional discrete diffusion models, \abbr~uses BERT as its backbone to perform text generation. The main differences are highlighted in {\color{magenta}{color}}: (1) \abbr~performs decoding without knowing the current time step while canonical diffusion models are conditioned on time step. (2) The diffusion process of \abbr~is non-Markovian in that it generates noise samples $\mathbf{x}_t$ conditioning not only on $\mathbf{x}_{t-1}$ but also on $\mathbf{x}_0$. Such a non-Markov process is due to our proposed noise schedule.}
    \label{fig: overview}
\end{figure}

\section{Introduction}
Diffusion models~\cite{sohl2015deep,ho2020DDPM,song2020DDIM} have recently emerged as a new class of state-of-the-art generative models, achieving high-quality synthesis results on image data~\cite{ramesh2022DALLE2,Rombach2022LDM,Saharia2022Imagen}. Though these models captured widespread attention from not only the research community but also the public, applying diffusion models to text data is still challenging and under-explored due to the discrete nature of the text. A few prior works that explored using diffusion models on text data can be divided into two lines. The first is to extend diffusion models to discrete state spaces~\cite{hoogeboom2021argmax,austin2021structured}. The second is to perform the diffusion process and its reverse process in the continuous domain and bridge the continuous and the discrete domain through embedding and rounding~\cite{li2022diffusion,Gong2022DiffuSeq}. However, none of these works leveraged pre-trained language models (PLMs,~\citet{Devlin2019BERT,Lewis2020BART,Raffel2020T5,brown2020gpt3,Qiu2020survey}), which are an unmissable treasure in the NLP community.

This work, to our knowledge, is the first attempt to combine diffusion models with PLMs. Such a combination is built upon a shared training objective between diffusion models and PLMs, i.e., \textit{denoising}. Diffusion models consist of a forward process (data to noise) and a reverse process (noise to data). In the forward process, a small amount of noise is gradually added to the data. Then, a neural network ($p_\theta$ in Figure~\ref{fig: overview}) is employed to learn the reverse process step by step, i.e., learn to denoise. Such a denoising neural network is naturally related to a wide class of PLMs that are pre-trained with denoising objectives such as BERT~\cite{Devlin2019BERT} and BART~\cite{Lewis2020BART}. Hence, pre-trained denoising language models can serve as a good start point to learn the reverse diffusion process. On the other hand, diffusion models also offer a promising training strategy for generative PLMs.
In contrast to commonly used generative PLMs (e.g., GPT~\cite{brown2020gpt3}) that relies on an autoregressive factorization of the joint probability, diffusion models provide another way of factorization along the dimension of time and therefore allow the model to be not necessarily autoregressive. Thus, diffusion models can be combined with a variety of PLMs that may not be pre-trained for generation.

In the discrete domain, the forward diffusion process can be implemented by a chain of transition matrices that gradually corrupt the clean text. As shown in Figure~\ref{fig: overview}, the clean text \texttt{"Hello world !"} is gradually corrupted into \texttt{"[MASK] [MASK] [MASK]"} during the diffusion process. In this work, we explore using pre-trained denoising language models (e.g., BERT) to learn the reverse diffusion process and demonstrate their advantages in accelerating convergence and improving generation quality. Further, we propose a new noise schedule of the forward process based on the principle of distributing the corrupted information uniformly across the forward process. The noise schedule, called \textit{spindle schedule}, generates noise for $\mathbf{x}_t$ conditioned not only on $\mathbf{x}_{t-1}$ but also on $\mathbf{x}_0$, making the forward process non-Markovian without changing the original training objective. Note that the denoising model takes as input $\mathbf{x}_t$ and time step $t$ to predict $\mathbf{x}_{t-1}$, where $t$ is unseen during the pre-training of language models so we investigate several ways of incorporating the time step into PLMs. As a result, we find that the best result is achieved by throwing away the time information, which we call \textit{time-agnostic decoding} (TAD).

Experimental results on unconditional text generation demonstrate the benefit of combining diffusion models with PLMs: the proposed \abbr~significantly improves the generation quality over existing diffusion models for text generation (e.g., D3PM~\cite{austin2021structured} and Diffusion-LM~\cite{li2022diffusion}) and previous generative masked language models (e.g., BERT-Mouth~\cite{wang2019berthasamouth}).
The effectiveness of the proposed spindle schedule and time-agnostic decoding is confirmed by ablation studies. In a nutshell, \abbr~enjoys the best of both worlds.

\begin{table*}[t]
\centering
\resizebox{\linewidth}{!}{
\begin{tabular}{lll}
\toprule
\multirow{6}{*}{BERT-Mouth}     & $t=0$ & \texttt{[MASK] [MASK] [MASK] [MASK] [MASK] [MASK] [MASK] [MASK] [MASK] [MASK]}                                                                 \\
                          & $t=8$ &  \texttt{[MASK] \textbf{of} [MASK] \textbf{five} [MASK] \textbf{remain} [MASK] \textbf{in} [MASK] \textbf{.}}                                                                 \\
                          & $t=16$ &     \texttt{\textbf{two of} [MASK] \textbf{five structures remain} [MASK] \textbf{this location .}}                                                              \\
                          & $t=24$ &    \texttt{\textbf{five of} [MASK] \textbf{the windows remain at this location .}}                                                               \\
                          & $t=32$   & \texttt{\textbf{most of even the windows stand still this day .}} \\ \midrule
\multirow{6}{*}{D3PM} & $t=0$ &  \texttt{[MASK] [MASK] [MASK] [MASK] [MASK] [MASK] [MASK] [MASK] [MASK] [MASK]}                                                                                               \\
                          & $t=8$ &  \texttt{[MASK] [MASK] [MASK] [MASK] \textbf{been} [MASK] [MASK] [MASK] [MASK] \textbf{.}}                                                                 \\
                          & $t=16$ &    \texttt{[MASK] [MASK] [MASK] [MASK] \textbf{been} [MASK] [MASK] \textbf{the} [MASK] \textbf{.}}                                                              \\
                          & $t=24$ &    \texttt{[MASK] [MASK] [MASK] \textbf{also been} [MASK] \textbf{by the } \texttt{[MASK]} \textbf{ .}}                                                               \\
                          & $t=32$   &  \texttt{\textbf{the man has also been arrested by the police .}} \\ \midrule
\multirow{6}{*}{\abbr}      & $t=0$ & \texttt{[MASK] [MASK] [MASK] [MASK] [MASK] [MASK] [MASK] [MASK] [MASK] [MASK]}                                                                                                \\
                          & $t=8$ &    \texttt{[MASK] \textbf{,} [MASK] [MASK] [MASK] [MASK] [MASK] \textbf{that} [MASK] .}                                                               \\
                          & $t=16$ &   \texttt{\textbf{today ,} [MASK] \textbf{will be} [MASK] [MASK] \textbf{that} [MASK] .}                                                                \\
                          & $t=24$ &   \texttt{\textbf{today ,} [MASK] \textbf{will be remembered for that mistake .}}                                                                \\
                          & $t=32$   & \texttt{\textbf{today , he will be remembered for that mistake .}}\\ \bottomrule
\end{tabular}
}
\caption{Examples generated by three generative masked language models showing the difference of noise schedule and generation quality.}
    \label{tab:generativeprocessofdb}
\end{table*}

\section{Background}
\label{sec: background}
\subsection{Diffusion Models}
Diffusion models~\citep{sohl2015deep, ho2020DDPM} are a class of latent variable models that are originally designed for continuous domains. A diffusion model is consisting of a forward diffusion process and a reverse diffusion process. Given a sample $\mathbf{x}_0\sim q(\mathbf{x}_0)$, a Markov chain of latent variables $\mathbf{x}_1,\cdots,\mathbf{x}_T$ are produced in the forward process by progressively adding a small amount of Gaussian noise to the sample:
\begin{equation}
    q(\mathbf{x}_t\vert\mathbf{x}_{t-1})=\mathcal{N}(\mathbf{x}_t;\sqrt{1-\beta_t}\mathbf{x}_{t-1},\beta_t\mathbf{I}),
\end{equation}
where $\{\beta_t\in(0,1)\}_{t=1}^T$ is a noise schedule controlling the step size of adding noise. Eventually $\mathbf{x}_T$ becomes an isotropic Gaussian distribution. If $\beta_t$ is small enough, the reverse process $q(\mathbf{x}_{t-1}\vert \mathbf{x}_t)$ is also a Gaussian, which is learned by a parameterized model
\begin{equation}
    p_\theta(\mathbf{x}_{t-1}\vert \mathbf{x}_t, t)=\mathcal{N}(\mathbf{x}_{t-1};\mu_\theta(\mathbf{x}_t,t),\Sigma_\theta(\mathbf{x}_t,t)),
\end{equation}
where $\mu_\theta(\cdot)$ and $\Sigma_\theta(\cdot)$ can be implemented by a U-Net or a Transformer. When conditioning also on $\mathbf{x}_0$,  $q(\mathbf{x}_{t-1}\vert \mathbf{x}_t, \mathbf{x}_0)$ has a closed form so we can manage to minimize the variational lower bound to optimize $\log p_\theta(\mathbf{x}_0)$:
\begin{align}
\label{equation: vlb}
    & \mathcal{L}_{\text{vlb}} = \mathbb{E}_{q} [D_\text{KL}(q(\mathbf{x}_T \vert \mathbf{x}_0) \parallel p_\theta(\mathbf{x}_T))] \nonumber \\
    & + \mathbb{E}_{q} [\sum_{t=2}^T D_\text{KL}(q(\mathbf{x}_{t-1} \vert \mathbf{x}_t, \mathbf{x}_0) \parallel p_\theta(\mathbf{x}_{t-1} \vert\mathbf{x}_t, t))] \nonumber \\
    & - \log p_\theta(\mathbf{x}_0 \vert \mathbf{x}_1),
\end{align}
where $\mathbb{E}_{q}(\cdot)$ denotes the expectation over the joint distribution $q(\mathbf{x}_{0:T})$.

\subsection{Diffusion Models in Discrete Domain}
For discrete domains, each element of $\mathbf{x}_t$ is a  discrete random variables with $K$ categories. For text data, $K=\vert V\vert$ is the size of the vocabulary. Denote $\mathbf{x}_t$ as a stack of one-hot vectors, the process of adding noise can be written as
\begin{equation}
    q(\mathbf{x}_t\vert \mathbf{x}_{t-1})=\texttt{Cat}(\mathbf{x}_{t};\mathbf{p}=\mathbf{x}_{t-1}\mathbf{Q}_t),
\end{equation}
where $\texttt{Cat}(\cdot)$ is a category distribution and $\mathbf{Q}_t$ is a transition matrix that is applied to each token in the sequence independently: $[\mathbf{Q}_t]_{i,j}=q(x_t=j\vert x_{t-1}=i)$. It is easy to obtain that
\begin{align}
\label{equation: cat_q}
    q(\mathbf{x}_{t-1}\vert & \mathbf{x}_t,\mathbf{x}_0)= \frac{q(\mathbf{x}_t \vert \mathbf{x}_{t-1}, \mathbf{x}_0) q(\mathbf{x}_{t-1} \vert \mathbf{x}_0) }{ q(\mathbf{x}_t \vert \mathbf{x}_0)} \nonumber\\ 
    =\texttt{Cat}&\left (\mathbf{x}_{t-1};\mathbf{p}=\frac{\mathbf{x}_t\mathbf{Q}_t^\top\odot\mathbf{x}_0\overline{\mathbf{Q}}_{t-1}}{\mathbf{x}_0\overline{\mathbf{Q}}_t\mathbf{x}_t^\top} \right),
\end{align}
where $\overline{\mathbf{Q}}_t=\mathbf{Q}_1\mathbf{Q}_2\cdots\mathbf{Q}_t$. Note that $\odot$ is element-wise multiplication and the division is row-wise. 

With $q(\mathbf{x}_{t-1}\vert  \mathbf{x}_t,\mathbf{x}_0)$ at hand, according to Eq. (\ref{equation: vlb}), we can use a parameterized model $p_\theta(\mathbf{x}_{t-1}\vert\mathbf{x}_t, t)$ to learn the reverse diffusion process.

\section{\abbr}
In contrast to recently proposed diffusion models for text, e.g., Diffusion-LM~\cite{li2022diffusion} and DiffuSeq~\cite{Gong2022DiffuSeq}, which are based on \textit{continuous} diffusion models, we instead explore \textit{discrete} diffusion models to integrate PLMs as the backbone. We first introduce a specific instance of discrete diffusion models~\cite{austin2021structured}, which considers a transition matrix with an absorbing state for the sake of using PLMs (\S~\ref{sec:absorbing}). Secondly, we introduce a new noise schedule of the forward diffusion process, called spindle schedule, which is based on the principle of distributing the corrupted information uniformly across the forward process (\S~\ref{sec: IIFP}). Then, we investigate several alternatives of incorporating the time step into PLMs for predicting $\mathbf{x}_{t-1}$ given $\mathbf{x}_t$ and $t$ (\S~\ref{sec: timesteps}).

\subsection{Diffusion Models with a Discrete Absorbing State}
\label{sec:absorbing}
To be combined with pre-trained denoising language models, we incorporate an \textit{absorbing state}, e.g., \texttt{[MASK]} for BERT, in the Markov process. In particular, each token in the sequence either stays the same or transitions to \texttt{[MASK]} with some probability. Formally, each entry of the transition matrix at step $t$ is as follows,
\begin{align}
    [\mathbf{Q}_t]_{i,j} = 
    \begin{cases}
        1 & \text{if}\  i = j = \texttt{[M]}, \\
        \beta_t & \text{if}\  j = \texttt{[M]}, i \ne \texttt{[M]}, \\
        1 - \beta_t & \text{if}\  i = j \ne \texttt{[M]},
    \end{cases}
\end{align}
where \texttt{[M]} is the abbreviation of \texttt{[MASK]}. Such a Markov process converges to a stationary distribution $q(\mathbf{x}_T)$, which places all probability mass on a sequence with all \texttt{[MASK]} tokens.



The $t$-step marginal $q(\mathbf{x}_t^i\vert\mathbf{x}_0^i)$ can be easily obtained in a closed form,
\begin{align}
\label{equation: qt given q0}
    q(\mathbf{x}_t^i\vert\mathbf{x}_0^i) =
    \begin{cases}
        \overline{\alpha}_t & \text{if}\  \mathbf{x}_t^i = \mathbf{x}_0^i, \\
        1 - \overline{\alpha}_t & \text{if}\ \mathbf{x}_t^i = \texttt{[M]},
    \end{cases}
\end{align}
where $\overline{\alpha}_t=\prod_{i=1}^t(1 - \beta_i)$,  $\mathbf{x}_t^i$ denotes the $i$-th token in the sequence at step $t$. Combining with Eq. (\ref{equation: vlb}) and (\ref{equation: cat_q}), we can derive a training objective to optimize $p_\theta(\mathbf{x_{t-1}\vert \mathbf{x}_t}, t)$ and generate a sample by performing the reverse diffusion process:
\begin{equation}
    p_\theta(\mathbf{x}_{0:T})=p(\mathbf{x}_T)\prod_{t=1}^Tp_\theta(\mathbf{x}_{t-1}\vert\mathbf{x}_t, t).
\end{equation}

\subsection{Spindle Noise Schedule}
\label{sec: IIFP}
The noise schedule in the continuous domain, such as the linear schedule~\cite{ho2020DDPM} and the cosine schedule~\cite{Nichol2021Improved}, has shown to be important to the performance of diffusion models. 

In contrast to the continuous domain where the noise can be easily controlled by the variance of the Gaussian, \textit{(1) it is less obvious how to control the degree of noise added at each step in the discrete domain}. For the discrete domain, the noise schedule $\beta_t=(T-t+1)^{-1}$ has been explored for the case of the uniform transition matrix~\cite{sohl2015deep,hoogeboom2021argmax} and the absorbing-state transition matrix~\cite{austin2021structured}. However, \textit{(2) such a schedule assumes all tokens carry the same amount of information and does not consider the linguistic difference among the tokens in a sequence}. Besides, \textit{(3) it violates the easy-first-generation nature of denoising language models}. That is, the model tends to generate tokens that are most frequently appearing (and is least surprising) in the training corpus to achieve a higher likelihood. As the context becomes richer, more details come up in the sequence. 

To address the above issues, we consider a noise schedule that (1) measures the added noise at each step by the corrupted information and encourage the corrupted information to be uniformly distributed across the diffusion steps. Since the information is measured independently for each token, (2) different tokens in a sequence are assigned different probabilities of transitioning to the \texttt{[MASK]} token. Moreover, inspired by the easy-first-generation phenomenon, (3) we put the tokens in a sequence in descending order of their information and divide them into $T$ buckets. Each bucket is ensured to contain the same amount of information. That is, we mask the most informative tokens at the start of the forward process and mask the least informative tokens at the end of the forward process such that the learnable reverse process follows an easy-first generative behavior.

In particular, distributing corrupted information uniformly across the forward steps can be formally described by
\begin{equation}
    1-\frac{t}{T}=\frac{\sum_{i=1}^nH(\mathbf{x}_t^i)}{\sum_{i=1}^nH(\mathbf{x}_0^i)}=\frac{\sum_{i=1}^n\overline{\alpha}_t^iH(\mathbf{x}_0^i)}{\sum_{i=1}^nH(\mathbf{x}_0^i)},
\end{equation}
where $H$ denotes the entropy, which measures the amount of information of a random variable, $\mathbf{x}^i$ denotes the $i$-th token in the sequence and $n$ denotes the length of the sequence. According to Eq. (\ref{equation: qt given q0}), $\overline{\alpha}_t^i=\prod_{j=1}^t(1-\beta_j^i)$ denotes the probability that the $i$-th token remains the same at step $t$, i.e., $\mathbf{x}_t^i=\mathbf{x}_0^i$. 
We expect that $\overline{\alpha}_t^i > \overline{\alpha}_t^j$ if $H(\mathbf{x}_t^i) < H(\mathbf{x}_t^j)$ such that easy (low-information) tokens emerges earlier than hard (high-information) tokens during the reverse process. 

\begin{figure}[t!]
    \centering
    \includegraphics[width=\linewidth]{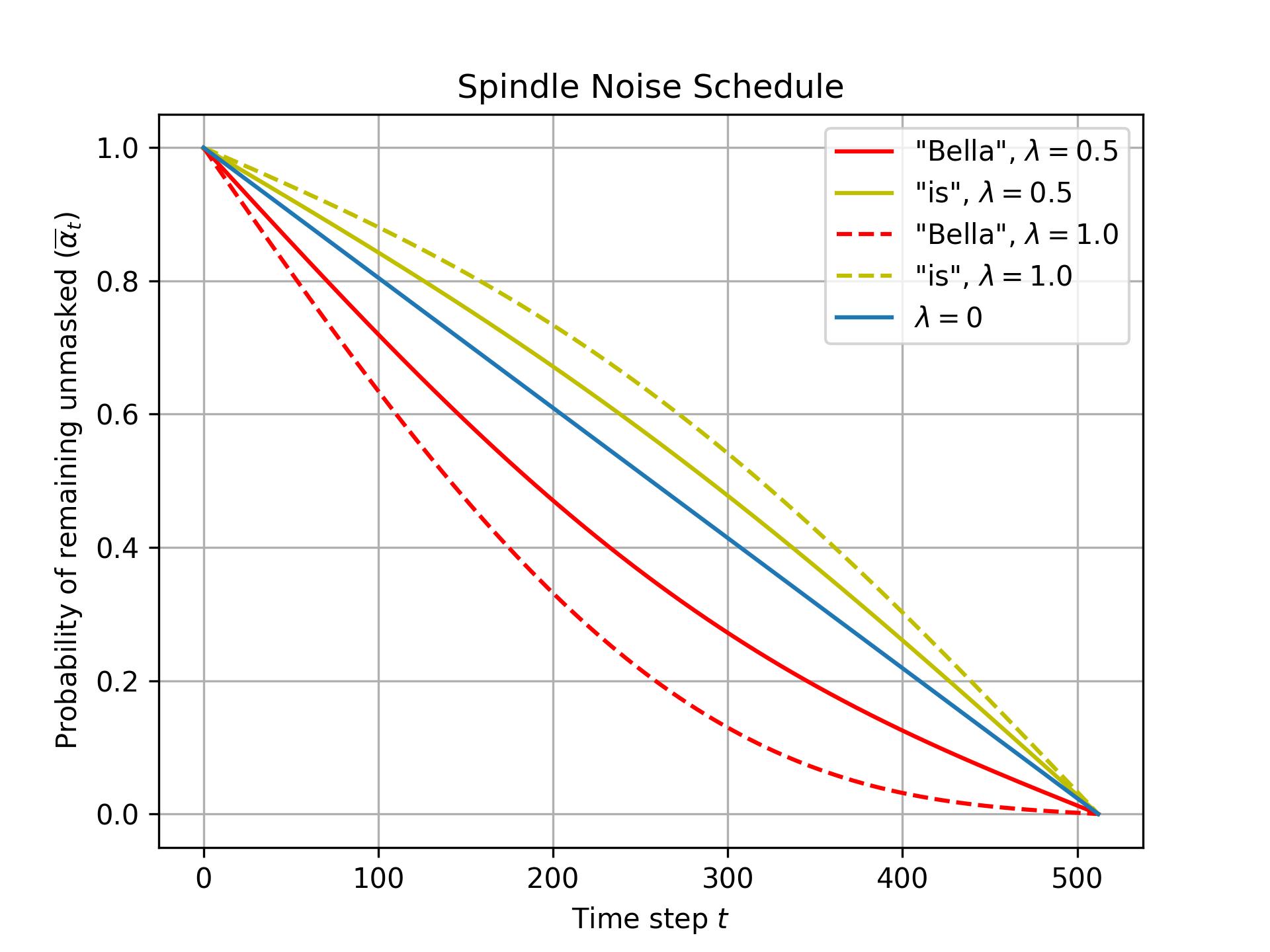}
    \caption{Each token in a sequence has a specific noise schedule depending on how much information is lost when they are masked. For instance, in the sentence "\texttt{Bella is sitting over there.}", "\texttt{Bella}" is the most informative word. Thus it is encouraged to be masked at the early stage so that our model learns to recover it in the last place.}
    \label{fig: word freq example}
\end{figure}

Considering these aforementioned properties, we construct $\overline{\alpha}_t^i$ as follows,
\begin{equation}
\label{equation: qt given q0 ours}
    \overline{\alpha}_t^i = 1 - \frac{t}{T} - S(t)\cdot \Tilde{H}(\mathbf{x}_0^i),
\end{equation}
\begin{equation}
    S(t) = \lambda\sin\frac{t\pi}{T},
\end{equation}
\begin{equation}
    \Tilde{H}(\mathbf{x}_0^i) = 1 - \frac{\sum_{j=1}^nH({\mathbf{x}_0^j})}{n H({\mathbf{x}_0^i})},
\end{equation}
where $S(t)$ is introduced to control the effect of the informativeness at time step $t$. It is designed to be sinusoidal to ensure $S(0)=S(T)=0$ such that $\mathbf{x}_t$ can retain all (zero) information when $t=0$ ($t=T$). The effect of $S(t)$ is controlled by a hyperparameter $\lambda$. When $\lambda=0$, the noise schedule is degraded to $\beta_t=(T-t+1)^{-1}$ as in \citet{sohl2015deep,hoogeboom2021argmax,austin2021structured}. Figure~\ref{fig: word freq example} shows how $\overline{\alpha}$ progresses during the forward process. The schedule is named as \textit{spindle} due to the shape of the probability curves.

In our proposed schedule, the transition probability at time step $t$ depends not only on the current state but also on the original text, making the forward diffusion process non-Markovian. Nevertheless, as revealed by Eq. (\ref{equation: cat_q}), this does not change the original training objective.

\subsection{The Design Space of Feeding Time Steps}
\label{sec: timesteps}
Typically, a diffusion model takes as input a noised sample and the time step to predict the denoised sample during the reverse process, i.e., $p_\theta(\mathbf{x}_{t-1}\vert \mathbf{x}_t, t)$. However, $t$ is an additional variable that is unseen during the pre-training of language models and therefore it is less trivial how to feed the time information into the PLMs. Here we explore three design choices of feeding time steps.


\paragraph{Layer-wise Time Embedding}
A straightforward choice is to include the time step as the same way as positional encoding, i.e., using the Transformer sinusoidal embedding or a learnable MLP in each Transformer layer. Note that this way is commonly adopted in previous work~\cite{ho2020DDPM,austin2021structured, li2022diffusion}.


\paragraph{Prefix Time Embedding}
Prompting language models by prepending trainable soft tokens to the input sequence has shown promising results recently~\cite{lester2021Prompt,sun2022BBT}. Hence, we also explore including a time step token embedding $\mathbf{v}(t)$ as a prefix of the input token embeddings $\langle\mathbf{v}(\mathbf{x}_t^1), \mathbf{v}(\mathbf{x}_t^2), \cdots, \mathbf{v}(\mathbf{x}_t^n)\rangle$. In particular, the time step token is inserted in between the \texttt{[CLS]} token and the input sequence. These added time step token embeddings are trained along with the PLM.

\paragraph{Time-Agnostic Decoding}
Another alternative is not to explicitly incorporate the time step $t$ because it can be implied by the noised sample $\mathbf{x}_t$.
In contrast to the image data, it is easier to implicitly infer the diffusion time step by counting the number of corrupted tokens (i.e., \texttt{[MASK]}) in the noised sequence. In this way, the PLM has to perform iterative decoding while being ignorant of the current time step, i.e., $p_\theta(\mathbf{x}_{t-1}\vert \mathbf{x}_t)$.


\section{Experiments}

\begin{table*}[t!]
\centering
\resizebox{\linewidth}{!}{
\begin{tabular}{@{}lcccccc@{}}
\toprule
\textbf{Method} & \textbf{Pretrained} & \textbf{Schedule} & \textbf{Time Step} & \textbf{PPL $\downarrow$} & \textbf{BLEU $\uparrow$} & \textbf{Self-BLEU $\downarrow$} \\ \midrule
\multirow{4}{*}{\begin{tabular}[c]{@{}c@{}}D3PM~\citep{austin2021structured}\end{tabular}} & \multirow{4}{*}{\XSolidBrush} & \multirow{2}{*}{$(T-t+1)^{-1}$} & LTE & 82.34 & 0.3897 & 0.2347 \\ \cmidrule(l){4-7} 
& & & \uwave{TAD} & 125.15 & 0.3390 & 0.2720 \\ \cmidrule(l){3-7} 
& & \uwave{Spindle} & LTE & \underline{77.50} & \underline{0.4241} & 0.2288 \\ \midrule
\multirow{2}{*}{\begin{tabular}[c]{@{}c@{}}Diffusion-LM~\citep{li2022diffusion}\end{tabular}} & \XSolidBrush & Cosine & LTE & 118.62 & 0.3553 & 0.2668 \\ \cmidrule(l){3-7} 
& \Checkmark & Cosine & LTE & 132.12 & 0.3562 & 0.2798 \\ \midrule 
BERT-Mouth~\citep{wang2019berthasamouth} & \Checkmark & - & - & 142.89 & 0.2867 & \textbf{0.1240} \\ \midrule
\multirow{5}{*}{\uwave{\abbr}}                                                                                        & \multirow{5}{*}{\Checkmark}   & \multirow{3}{*}{$(T-t+1)^{-1}$} & LTE & 92.53 & 0.3995 & 0.2118 \\ \cmidrule(l){4-7} 
& & & \uwave{PTE} & 79.95 & 0.3886 & 0.2156 \\ \cmidrule(l){4-7} 
& & & \uwave{TAD} & 78.76 & 0.4213 & \underline{0.2116} \\ \cmidrule(l){3-7} 
& & \uwave{Spindle} & \uwave{TAD} & \textbf{63.78} & \textbf{0.4358} & 0.2151 \\ \bottomrule
\end{tabular}}
\caption{Main results on LM1B. The methods proposed in this work are marked with \uwave{wavy lines}. The best results are in \textbf{bold} and the second best results are \underline{underlined}. LTE: layer-wise time embedding. PTE: prefix time embedding. TAD: time-agnostic decoding.}
\label{tab: main results}
\end{table*}

\subsection{Experimental Setup}
\label{sec: setup}
We mainly focus on unconditional text generation in complex scenarios where the training data covers a wide range of topics and is composed of a large vocabulary. Experiments are conducted on the One Billion Word dataset (LM1B) \citep{chelba2013lm1b}. LM1B is a language corpus with about 30 million sentences and a vocabulary of about 793k. We use the standard train-test split and take 1\% of the training set for validation. All text data are lower-cased to align with the settings of \citet{austin2021structured}.

Our \abbr~is based on \textsc{Bert-Base-Uncased} with about 110M parameters. We train \abbr~using the AdamW optimizer~\cite{loshchilov2019AdamW} for 1.9 million steps with learning rate of 3e-6, dropout probability of 0.1, batch size of 32. For the first 10K steps, we use a linear warmup schedule starting from learning rate of 1e-8. All experiments are conducted on NVIDIA A100 Tensor Core GPUs. We use 4 GPUs for training and a single GPU for sampling.

\subsection{Baselines}
We conduct comparison on unconditional text generation against several non-autoregressive (NAR) baselines: D3PM \citep{austin2021structured}, Diffusion-LM \citep{li2022diffusion}, and BERT-Mouth \citep{wang2019berthasamouth}.\footnote{Another strong baseline of NAR text generation is SUNDAE \citep{savinov2021sundae} but unfortunately there is no public implementation available. We will include comparison with SUNDAE in later versions by directly using the results reported in the original paper and use the same settings to train \abbr~for fair comparison.} 

\paragraph{D3PM}
D3PM is a general framework of discrete diffusion models. We implement an instance of D3PM with the absorbing state and a layer-wise time embedding. Both \abbr~and D3PM are implemented with a sequence length $n=128$ and diffusion steps $T=2048$. During inference, we perform decoding with 16 time steps in each iteration. The total inference cost is 128 iterations, which is smaller than that chosen in existing diffusion or diffusion-like models for unconditional generation \citep{hoogeboom2021argmax, savinov2021sundae}. This has no impact on our conclusions since increasing the diffusion step does not bring substantial improvement.

\paragraph{Diffusion-LM}
Diffusion-LM learns an embedding to map discrete text into the continuous space where it performs Gaussian diffusion process. A rounding step is required to map the continuous embeddings into discrete texts. We re-implemented Diffusion-LM with the model architecture of BERT and diffusion steps $T=2000$. Since the performance drop of Diffusion-LM is bigger than D3PM and \abbr~when we sample less steps during generation, we do not skip steps so the number of inference is about 4 times that of \abbr~and the exact generation time comparison is reported in \S~\ref{sec: efficiency}.

\paragraph{BERT-Mouth}
BERT-Mouth samples text from BERT via order-agnostic autoregressive masked language modeling. Starting from a sequence of \texttt{[MASK]}, BERT samples one token at each time step in random order. Another option is decoding from left to right, like autoregressive models. In our preliminary experiments, we find that random position sampling performs better. We continue pretraining BERT on LM1B to adapt BERT to downstream training corpus.

\subsection{Main Results}
Our main results are included in Table~\ref{tab: main results}. We choose BLEU-4 as the metric for generation quality and diversity. For each method, we sample 1K text for evaluating BLEU score and another 1K for self-BLEU. Note that with different sampling strategy, the BLEU/self-BLEU results may vary. For fair comparison, the sentences sampled by D3PM and \abbr~have a fixed length $n=64$ and are sampled by a top-$K$ filter where $K=30$. Diffusion-LM and BERT-Mouth are trained and sampled following their original implementation. Overall, \abbr~achieves the best generation quality and diversity trade-off among the considered NAR methods. Besides, the perplexity of \abbr~with the spindle noise schedule is substantially higher. Evidence of lower bound is used as a proxy of the perplexities of \abbr~and D3PM since the exact likelihood of diffusion models is intractable.

\paragraph{\abbr~vs. Other Generative BERT Models}
We compare \abbr~ with another representative generative masked language model, BERT-Mouth~\cite{wang2019berthasamouth}. 
Experimental results show that \abbr~achieves better performance in terms of the perplexity and the BLEU score.
We attribute the superior performance of \abbr~to its one-time sampling of all tokens, which helps \abbr~generate more coherent text, especially in a long range. Although such decoding may face the problem of multimodality \citep{gu2017NAT}, inappropriate phrases can be fixed in the upcoming diffusion steps. The probabilistic modeling offers more flexibility in that generated tokens with low probability are more likely to be masked and re-sampled. In BERT-Mouth, however, the tokens are fixed once sampled. \citet{wang2019berthasamouth} also proposed to continue masking and predicting tokens after the whole sequence is complete, revising the sentence for higher quality. But such randomness in the selection and replacement of tokens results in low inference speed.

\begin{figure}[t!]
	\centering
	\includegraphics[width=\linewidth]{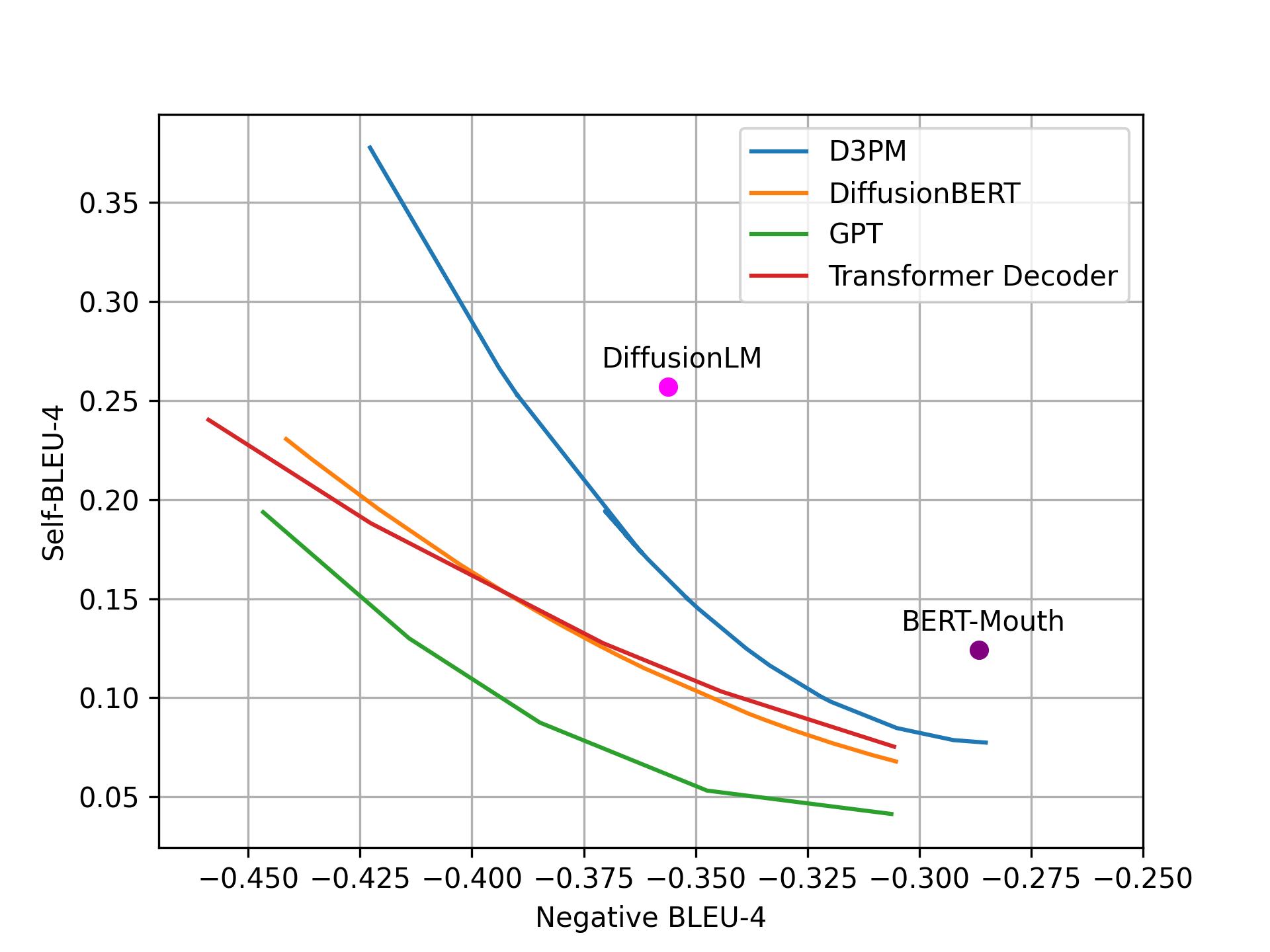}
	\caption{BLEU scores on the LM1B test set. Left is better, lower is better. For GPT and Transformer decoder, we control quality-variation with sampling temperature. D3PM and \abbr~are controlled by truncation sampling hyperparameter $K$.}
	\label{fig: q-d tradeoff}
\end{figure}

\paragraph{Discrete vs. Continuous Diffusion Models}
We then focus on the comparison of discrete and continuous diffusion models for text generation. 
To achieve this, we mainly compare \abbr~with recently proposed Diffusion-LM, which is based on continuous diffusion models.
As a result, despite of its outstanding controlling ability, we show that the texts generated by Diffusion-LM have a lower quality than \abbr.
Though both \abbr~and Diffusion-LM adopt the same configuration of Transformer, it is worth noting that the superior performance of \abbr~may be contributed by not only the discrete diffusion models but also the use of pre-trained models.
To disentangle the effect of pre-training and discrete/continuous diffusion models, we also explore initializing Diffusion-LM with BERT.
As shown in Table~\ref{tab: main results}, training Diffusion-LM from BERT initialization performs even worse than training from scratch.
We conjecture that the continuous nature of Diffusion-LM is not compatible with the initialization from BERT since the embedding learned by BERT may not be suitable for the Gaussian diffusion process.
In contrast, the comparison of D3PM and \abbr~shows that \abbr~benefits much from the BERT initialization due to its discrete diffusion process.
\begin{figure}[t!]
    \centering
    \includegraphics[width=\linewidth]{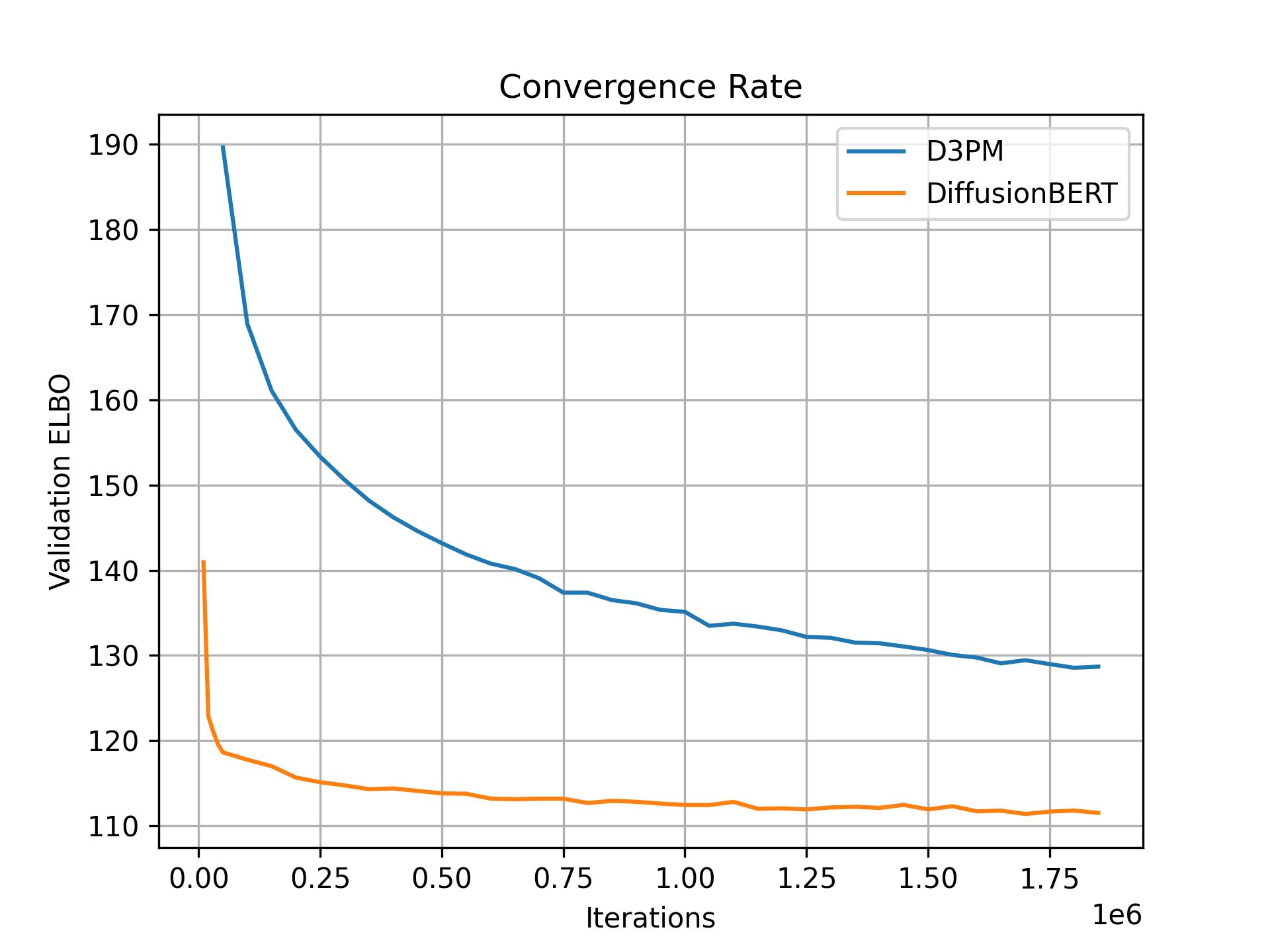}
    \caption{Curve of validation ELBO during training.}
    \label{fig: convergence}
\end{figure}
\paragraph{Effect of Time Step}
In terms of both likelihood and generation quality, the layer-wise time embedding (LTE) lags far behind the other two time step designs for \abbr~while time-agnostic decoding (TAD) achieves the best result. By contrast, D3PM without time step embedding performs significantly worse. In a nutshell, simplifying time step design has positive effect on \abbr~but is quite harmful for D3PM. This suggests that initializing $p_\theta$ with PLMs enables \abbr~to perform generation without explicitly providing time information yet achieving better generation results. 
The resemblance between BERT pre-training objective and absorbing diffusion models makes it easier for \abbr~to generalize to noisier scenarios while a Transformer encoder trained from scratch needs a specific time-aware module to model the reverse process.

\paragraph{Effect of the Spindle Noise Schedule}
We try our proposed spindle noise schedule on both \abbr~and D3PM. The perplexity is improved by 18\% and 19\% for D3PM and \abbr, respectively. Besides, D3PM with the spindle schedule outperforms that with the standard $(T-t+1)^{-1}$ schedule in generation quality. The same trend holds for \abbr~but with a smaller margin.

\subsection{Quality-Diversity Trade-off}
As shown in Figure \ref{fig: q-d tradeoff}, \abbr~exhibits comparable generation ability with a Transformer decoder trained from scratch and pushes the Pareto front of NAR generation quality/diversity trade-off by a large margin. However, it still falls behind pretrained AR models of the same size.
\begin{table}[t!]
\centering
\resizebox{\linewidth}{!}{\begin{tabular}{@{}cccc@{}}
\toprule
\textbf{Method}                       & \textbf{Steps} & \textbf{Inference Time (secs)} & \textbf{PPL} \\ \midrule
\multirow{6}{*}{\abbr}                & 2                        & 0.66                 & 313.57       \\
                                      & 8                        & 1.39                 & 91.01        \\
                                      & 16                       & 1.80                 & 75.66        \\
                                      & 64                       & 4.25                 & 65.83        \\
                                      & 128                      & 7.53                 & 63.78        \\
                                      & 512                      & 27.48                & 54.63        \\ \midrule
Diffusion-LM                           & 2000                     & 83.67                & 112.12       \\ \midrule
\multirow{2}{*}{BERT-Mouth} & 64                       & 2.18                   & 142.89           \\
                                      & 512                      & 14.39                   & 86.78           \\ \midrule
GPT                                   & 64                       & 1.55                 & 38.7         \\ \bottomrule
\end{tabular}}
\caption{Comparison of inference time and perplexity among baselines and \abbr.}
\label{tab: sampling speed}
\end{table}

\subsection{Efficiency of Training and Generation}
\label{sec: efficiency}
One important feature of \abbr~is that with time-agnostic decoding, all parameters are initialized by pretrained models. Consequently, \abbr~includes fewer parameters and is free from adapting new parameters, improving training and decoding efficiency.

\paragraph{Faster Convergence}
\abbr~converges remarkably faster than D3PM. Figure \ref{fig: convergence} demonstrates the curve of validation ELBO in the training process. Even if the training budget is cut to 30\% (i.e. 0.5 million steps), \abbr~is still able to match the performance reported in Table \ref{tab: main results}.

\paragraph{Sampling Speed}
With the $x_0$-parameterization proposed in \citet{song2020DDIM} and \citet{ austin2021structured}, \abbr~is able to perform inference with any given budget by controlling the step size in the reverse process. We also control the sampling time of BERT-Mouth by adjusting the max iteration count of its mask-predict process. We list the decoding speed and the corresponding perplexity on the LM1B test set in Table \ref{tab: sampling speed}. Overall, \abbr~exhibits competitive performance even when it reaches comparable speed to GPT and outperforms BERT-Mouth in efficiency-performance tradeoff.

\section{Related Work}

\subsection{BERT for Text Generation}
It has been shown by \citet{wang2019berthasamouth} that the transfer-learning ability of BERT does not only helps to achieve impressive results in natural language understanding but also benefits sequential sampling for text generation. However, its bi-directionality nature holds BERT from matching the decoder-only counterparts \citep{radford2018gpt} in modeling text from left to right.

\subsection{Diffusion Models for Text}
This work lies in the line of diffusion models, a latent variable generative framework proposed by \citet{sohl2015deep}. It has been architecturally improved by \citet{ho2020DDPM} and has gained broad attention for its impressive generation ability in continuous domain (e. g. image and audio) \citep{ramesh2022DALLE2, kong2020diffwave, nichol2021IDDPM}. 
Despite their great success and state-of-the-art sample quality in the above domains, diffusion models for text still struggle to match autoregressive models in various generation tasks. Since the Gaussian noise proposed in \citet{sohl2015deep} cannot be directly applied to discrete data, they also introduced a discrete forward process with a Bernoulli transition kernel. \citet{hoogeboom2021argmax} made a step forward from Bernoulli to categorical distributions. A more general family of discrete diffusion processes was introduced in \citet{austin2021structured, hoogeboom2021OA-ARDM}, including absorbing kernels and combinations of absorbing and uniform transition kernels.
\citet{li2022diffusion} models text in the continuous embedding space, which is closer to the settings in earlier works of diffusion models and shows impressive performance in classifier-controlled text generation. While the decoding and convergence speed are substantially slower and the generated text lacks coherence. Moreover, in scenarios where the vocabulary is large, the k-nearest-neighbor algorithm used in decoding holds up decoding even more severely.

\subsection{Non-Autoregressive Text Generation}
Absorbing discrete diffusion models resembles conditional masked language models (CMLMs) \citep{ghazvininejad2019CMLM} in that both methods predict the whole sequence simultaneously and follows a construct-destruct pattern to iteratively refine the generated text. The main difference lies in the training objective: \abbr~models a stochastic process and drives BERT to learn a group of distributions to gradually recover training data while CMLMs forces the neural network to deterministically recover the whole sequence in every iteration, thus it fails to explicitly model the denoising process. \citet{savinov2021sundae} proposed to approach the problem of non-autoregressive text modeling via unrolling the generation path to prepare the model for the partially corrupted sequences it will encounter during generation, which resembles the idea of diffusion models for unconditional text generation. Non-autoregressive models are also considered in translation but implemented in various ways, e.g., insertion/deletion \citep{gu2019levenshtein} and iterative sequence alignment \citep{saharia2020NATlatentalignments}.

\section{Conclusion}
This work aims to approach the problem of unconditional text generation for non-autoregressive models. To achieve this, we combine pretrained language models with absorbing-state discrete diffusion models for text. The training procedure of our proposed \abbr~includes two main deviations from current discrete diffusion models, i.e., a new family of time step designs and the spindle noise schedule. The novel spindle noise assigns a schedule for each token according to its frequency in the training corpus. Experimental results demonstrate the success of \abbr~in terms of perplexity. It also pushes the Pareto front of quality-variance tradeoff of NAR methods by a large margin, comparable to a Transformer decoder trained from scratch.

\bibliography{anthology,custom}
\bibliographystyle{acl_natbib}




\end{document}